\documentclass{article}

\usepackage[preprint,nonatbib]{neurips_2024}

\usepackage[utf8]{inputenc} 
\usepackage[T1]{fontenc}    
\usepackage[pagebackref, breaklinks=true, colorlinks,citecolor=blue, linkcolor=red, bookmarks=false]{hyperref}
\usepackage{url}            
\usepackage{booktabs}       
\usepackage{amsfonts}       
\usepackage{nicefrac}       
\usepackage{microtype}      
\usepackage{xcolor}         

\usepackage{subfig}
\usepackage{graphicx}
\usepackage{multirow}

\usepackage{amsthm}
\usepackage{amsmath}
\usepackage{wrapfig}        
\usepackage{colortbl}

\title{Orchestrating Dual-Boundaries: An Arithmetic Intensity Inspired Acceleration Framework for Diffusion Language Models}

\author{%
  \textbf{Linye Wei\quad Wenjue Chen\quad Pingzhi Tang\quad Xiaotian Guo\quad Le Ye\quad Runsheng Wang\quad Meng Li$^{\dag}$}\\
  Peking University\\
  \texttt{meng.li@pku.edu.cn} \\
  Codes:~\, \url{https://github.com/PKU-SEC-Lab/ODB-dLLM}
}

\begin{document}

\newcommand{\method}{ODB-dLLM}
\newcommand{\figref}[1]{Fig.~\ref{#1}}
\newcommand{\tabref}[1]{Tab.~\ref{#1}}
\newcommand{\secref}[1]{Sec.~\ref{#1}}

\maketitle

\begin{abstract}
Diffusion-based large language models (dLLMs) have recently gained significant attention for their exceptional performance and inherent potential for parallel decoding. Existing frameworks further enhance its inference efficiency by enabling KV caching. However, its bidirectional attention mechanism necessitates periodic cache refreshes that interleave prefill and decoding phases, both contributing substantial inference cost and constraining achievable speedup. Inspired by the heterogeneous arithmetic intensity of the prefill and decoding phases, we propose~\method, a framework that orchestrates dual-boundaries to accelerate dLLM inference.
In the prefill phase, we find that the predefined fixed response length introduces heavy yet redundant computational overhead, which affects efficiency. To alleviate this, \method~incorporates an adaptive length prediction mechanism that progressively reduces prefill overhead and unnecessary computation. In the decoding phase, we analyze the computational characteristics of dLLMs and propose a dLLM-specific jump-share speculative decoding method to enhance efficiency by reducing the number of decoding iterations.
Experimental results demonstrate that~\method~achieves 46–162× and 2.63–6.30× speedups over the baseline dLLM and Fast-dLLM, respectively, while simultaneously mitigating the accuracy degradation in existing acceleration frameworks.
\end{abstract}

\section{Introduction}
\label{sec:introduction}

The rapid breakthroughs of large language models (LLMs) \cite{yang2025qwen3,grattafiori2024llama,liu2024deepseek} have ushered artificial intelligence into a new era. Their impressive capabilities have been widely demonstrated across diverse tasks \cite{jiang2024survey,liu2025aligning,yan2025entropy,yan2026pixel}. Nevertheless, the inherent limitations of the autoregressive decoding paradigm and causal attention mechanism limit decoding efficiency and hinder their applicability to tasks involving infilling or structured generation \cite{berglund2023reversal,dong2024xgrammar}.

Recently, diffusion-based large language models (dLLMs) \cite{nie2025large,ye2025dream,khanna2025mercury} have emerged, employing global decoding and bidirectional attention to enable native parallel decoding and controllable generation \cite{kim2025train,xiong2025unveiling}. Although these models achieve competitive performance with autoregressive models on standard tasks \cite{song2025seed,zhu2025llada15}, their bidirectional attention disables the use of KV cache, forcing each step to re-execute the entire prefill phase \cite{zhu2025lladamoe,wei2025accelerating}. This results in a severe compute-bound bottleneck, limiting the expected inference speed advantage of dLLMs.

Fast-dLLM \cite{wu2025fast} and similar approaches \cite{ben2025accelerated,chen2025dpad,jiang2025d} offer a solution to this challenge by partitioning a predefined-length response into multiple blocks. For each block, the model precomputes and caches the KV pairs of tokens outside the block, then performs decoding only within the block, accepting multiple tokens based on a confidence threshold. Once all positions in the block have been filled, the global KV cache is recomputed for the next block. This interleaved prefill–decoding scheme bridges dLLM’s parallel decoding and KV caching, and has rapidly become a widely adopted inference framework in the community.
  
However, several limitations remain in it. Although block-wise cache refresh removes the need for per-step prefill, it still introduces considerable inference cost, resulting in a compute–memory alternation bottleneck. The predefined-length response is also suboptimal, leading t  o high prefill overhead and redundant inference. Furthermore, stable KV approximation and task-agnostic fixed response lengths inevitably degrade model performance.

Our analysis of the interleaved compute/memory boundaries reveals two key limitations. On the one hand, the predefined response length represents a compromise under uncertainty rather than an optimal trade-off between accuracy and efficiency \cite{li2025diffusion,li2025beyond}. The full-sequence prefilling and unmasking processes reveal an opportunity to reduce computational load through length prediction. On the other hand, although the subsequent decoding phase involves all tokens within a block, its arithmetic intensity remains insufficient to overcome the memory wall of mainstream GPUs, leaving substantial headroom for further parallelization.

\begin{wrapfigure}{r}{0.62\textwidth}
    \centering
    \includegraphics[width=0.62\textwidth]{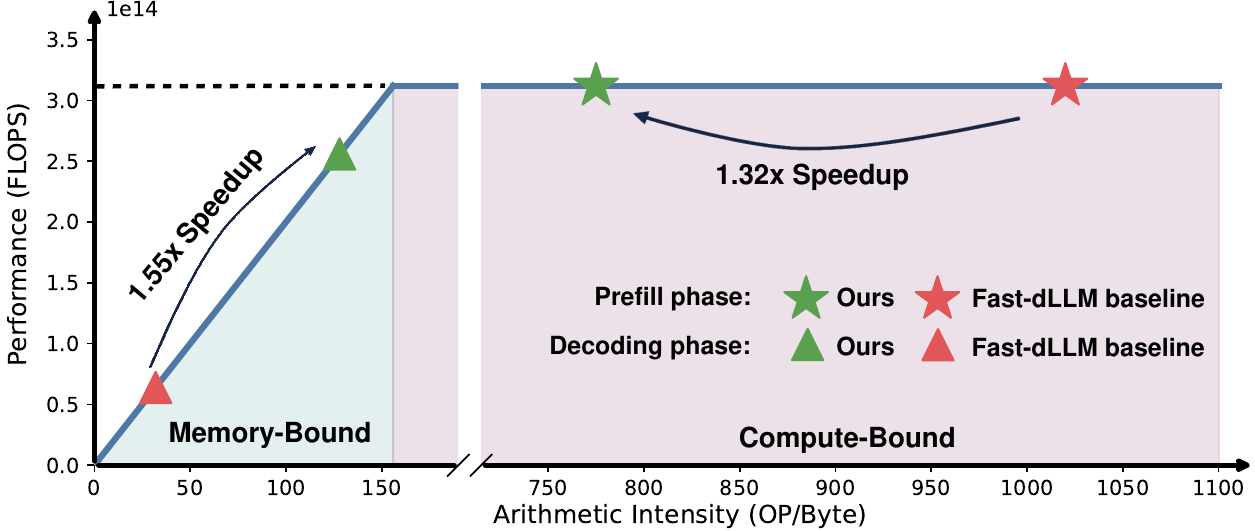}
    \caption{Inference with~\method~on GSM8K dataset \cite{cobbe2021training}. Comparison of the arithmetic intensity and speedup against prior framework on the roofline model of NVIDIA A100 GPU.}
    \label{fig:1}
\end{wrapfigure}

Motivated by these observations, we propose~\method, a training-free framework that orchestrates dual-boundaries for efficient dLLM inference, as illustrated in~\figref{fig:1}. 


By incorporating an adaptive length prediction strategy into each cache refresh, \method~progressively reduces the prefill overhead and eliminates redundant inference. 
Furthermore, we propose a jump-share speculative decoding tailored for dLLMs, which aligns with their block-wise parallel decoding by employing inter-block jump verification and sharing the KV pairs of decoded tokens.
By introducing additional computation during the memory-bound phases, this strategy increases the number of tokens accepted per iteration and effectively reduces the total decoding steps. 
Notably, \method~also achieves improved accuracy compared to existing frameworks.
Our main contributions are summarized as follows:
\begin{itemize}
    \item We identify the dual-boundary bottleneck in current dLLM acceleration frameworks, where periodic cache refreshes lead to alternating compute- and memory-bound phases.
    \item We propose~\method, which integrates adaptive length prediction and jump-share speculative strategy, optimizing the compute-memory characteristics of dLLM inference.
    \item Extensive experiments demonstrate that~\method~achieves 46-162× and 2.63-6.30× speedups over the baseline dLLM and Fast-dLLM, respectively, while alleviating the performance degradation in Fast-dLLM.
\end{itemize}

\section{Background}
\label{sec:background}

\subsection{Diffusion Language Models and Acceleration Techniques}
\label{bacground1}

The dominant paradigm for diffusion-based large language models (dLLMs) is the \textbf{Masked Diffusion Model (MDM)} \cite{sahoo2024simple,shi2024simplified,yan2026less}, utilizing a simple forward noising process where tokens are progressively replaced by a special token $\texttt{[MASK]}$ through time $t \in [0, 1]$. 
During generation, a $\tau$-leaping approximation~\cite{gillespie2001approximate} is commonly employed, which enables an iterative generation process, moving from a noise level $t$ to an earlier level $s < t$ by unmasking multiple tokens in a single step. 
Let $\boldsymbol{x}_t$ and $\boldsymbol{x}_t^i$ denote the discrete token sequences at time steps $t$ and the specific token at the $i$-th position, respectively.
The reverse transition probability $q_{s|t}$ is defined as:
\begin{align}
\label{eq:reverse_process}
q_{s|t}(\boldsymbol{x}_s^i|\boldsymbol{x}_t)=\begin{cases}
1, & \boldsymbol{x}_t^i \neq \texttt{[MASK]}, \boldsymbol{x}_s^i = \boldsymbol{x}_t^i \\
\frac{s}{t}, & \boldsymbol{x}_t^i = \texttt{[MASK]}, \boldsymbol{x}_s^i = \texttt{[MASK]} \\
\frac{t-s}{t} q_{0|t}(\boldsymbol{x}_s^i | \boldsymbol{x}_t), & \boldsymbol{x}_t^i = \texttt{[MASK]}, \boldsymbol{x}_s^i \neq \texttt{[MASK]}.
\end{cases}
\end{align}

\begin{figure}[th]
    \centering
    \includegraphics[width=0.75\linewidth]{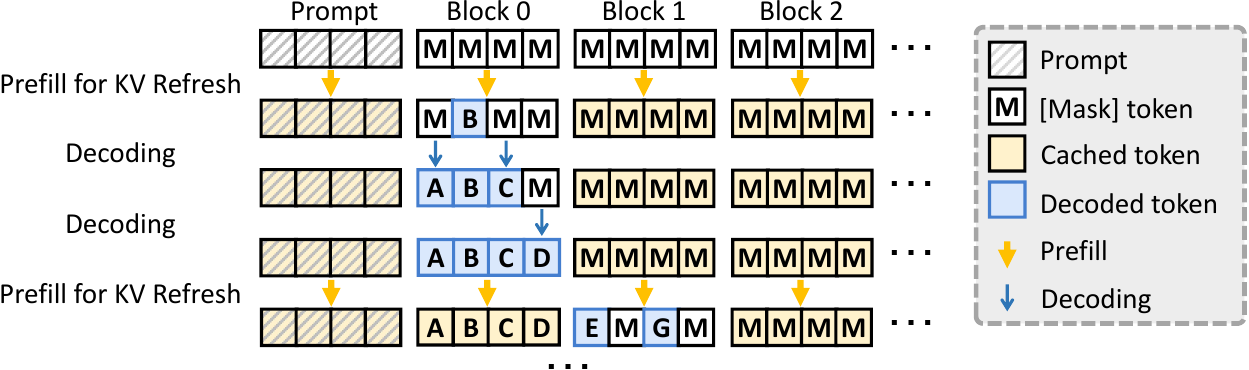}
    \caption{Inference with parallel decoding and DualCache.}
    \label{fig:2}
\end{figure}

A primary challenge in dLLM inference stems from its bidirectional attention, which prevents the use of standard Key-Value (KV) caching~\cite{pope2023efficiently} and leads to a severe compute-bound prefill bottleneck.
To mitigate this, \textbf{Fast-dLLM}~\cite{wu2025fast} introduced a widely adopted framework that partitions the response into blocks. 
As shown in~\figref{fig:2}, for each block, Fast-dLLM precomputes and caches the KV pairs for all tokens \textit{outside} the current block using a "DualCache" (caching both the prefix prompt and the suffix $\texttt{[MASK]}$ tokens). 
Once all positions in the block are filled, the global KV cache is refreshed for the next block.

However, this block-wise approach relies on a predefined fixed length, which causes a trade-off between performance and redundant computation.
To address this, dLLM-Var~\cite{yang2025diffusion} proposes to train the model to explicitly and accurately predict an $\texttt{[EOS]}$ (end-of-sequence) token, allowing it to natively determine the sequence boundary. 
Although effective, this method requires a dedicated training phase, hindering its straightforward application by end-users.
DAEDAL~\cite{li2025beyond} offers a training-free strategy using a complex two-phase process: iteratively expanding a coarse length \textit{before} denoising, and then dynamically inserting masks \textit{during} the denoising process. 
While flexible, this multi-stage heuristic can be less precise in predicting the optimal length early, which is crucial for minimizing prefill computation. 
Our work seeks a more direct and accurate training-free method to optimize the generation length dynamically during each cache refresh.

\subsection{Speculative Decoding in LLM/dLLM}
\label{sec:background2}

Speculative decoding is a widely adopted technique to accelerate autoregressive (AR) LLM inference~\cite{miao2023specinfer,chen2023accelerating,leviathan2023fast,cai2024medusa}. 
It employs a "Draft-then-Verify" paradigm, where a lightweight, efficient "draft" model first generates a sequence of candidate tokens autoregressively. 
These candidates are then verified in parallel by the large "target" model in a single forward pass. 
The process guarantees a lossless speedup identical to the target model's output, as it accepts tokens only up to the first mismatch. 

Applying this concept to dLLMs is not straightforward due to their non-autoregressive, bidirectional, and parallel-unmasking nature. 
Accordingly, speculative strategies for dLLMs must operate in a \textbf{block-wise} manner, aligning with the interleaved prefill-decoding process of frameworks like Fast-dLLM.
Given this, most approaches leverage the dLLM itself as its own drafter in an "auto-speculative" or "self-speculative" manner, avoiding the overhead of a separate draft model. 
SSD~\cite{gao2025self} uses the dLLM to generate and verify predictions for multiple positions simultaneously, organizing them into hierarchical verification trees to achieve lossless speedup. 
Similarly, Spiffy~\cite{agrawal2025spiffy} proposes a "directed draft graph" to structure the candidate states generated by the dLLM itself, which can then be verified in parallel.
This approach is designed to improve the quality-speed trade-off rather than guarantee lossless output.
While these methods effectively accelerate the memory-bound decoding phase, they share a common limitation: when a drafted block is rejected, the computational effort and the information contained within that rejected draft are largely discarded. 
Our work addresses this inefficiency by proposing a \textbf{jump-share} speculative strategy that, unlike previous methods, utilizes information from unaccepted blocks to inform subsequent generation steps, thereby further optimizing the compute-memory trade-off.
\tabref{tab1} presents a comparison between our framework and existing LLM/dLLM with their speculative approaches.

\begin{table}[!th]
\caption{Comparison between~\method~and autoregressive/diffusion LLMs (ARMs/DLMs) with their speculative methods (last three columns).}
\centering
\small
\begin{tabular}{c|cc|ccc}
\toprule
Frameworks & Parallelism & Throughput & Draft Model & Draft Efficiency & Verify Efficiency\\
\midrule
ARMs & Low & Low & Necessary & High & Low \\
DLMs & Medium & Medium & Unnecessary & Low & High  \\
\rowcolor{gray!20}
Ours & High & High & Unnecessary & High & High  \\
\bottomrule
\end{tabular}
\label{tab1}
\end{table}
\section{Motivation}
\label{sec:motivation}



\begin{figure}[th]
    \centering
    \includegraphics[width=0.95\linewidth]{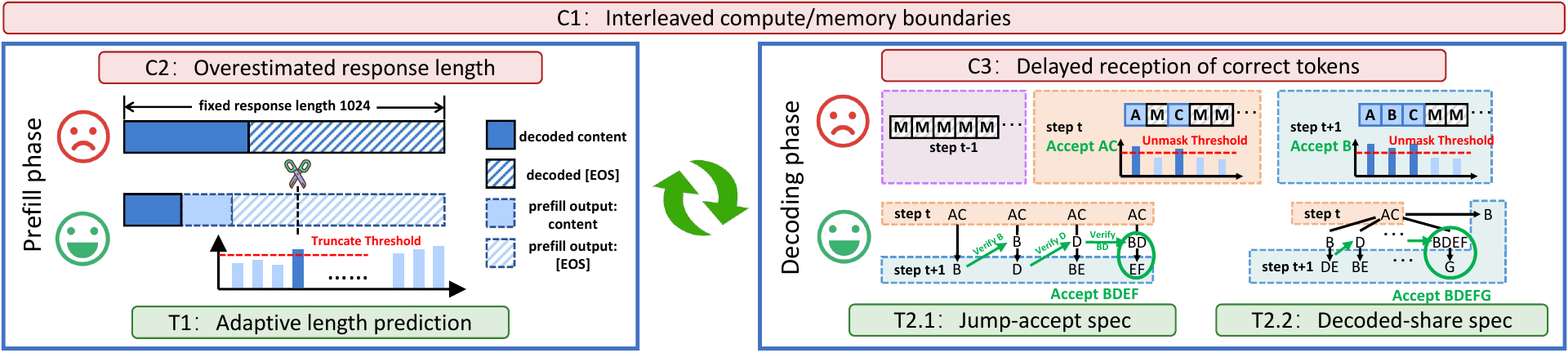}
    \caption{Dual-boundary challenges for dLLM and the overview of \method's design.}
    \label{fig:3}
\end{figure}

We profiled the alternating prefill and decoding phases—inherent to KV cache refreshes in existing dLLM frameworks—across multiple datasets. Our analysis reveals distinct variations in arithmetic intensity between these phases. As illustrated in~\figref{fig:3}, this discrepancy highlights significant inefficiencies and potential optimization opportunities. From this analysis, we derive three key observations that motivate our subsequent work.

\begin{figure}[th]
    \centering
    \includegraphics[width=0.8\linewidth]{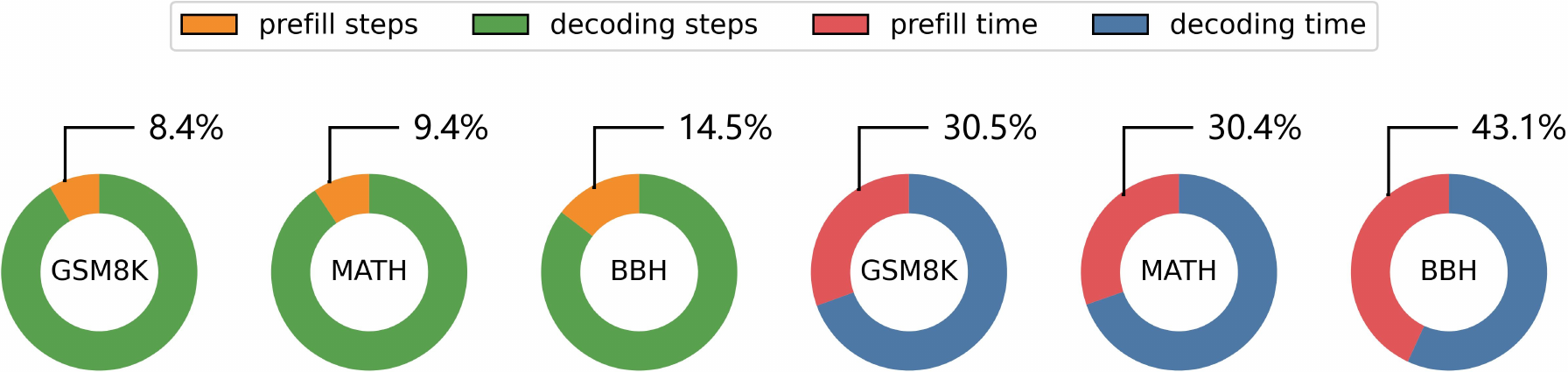}
    \caption{Proportion of step counts and execution time across prefill and decoding phases.}
    \label{fig:4}
\end{figure}

\textbf{Observation 1: The interleaved compute- and memory-bound patterns both constrains efficiency.} As discussed before, the response sequence is partitioned into blocks and decoded in a semi-autoregressive manner, where each new block triggers a global prefill to refresh the KV cache. As shown in~\figref{fig:4}, our profiling across GSM8K, Minerva Math, and BBH reveals that the prefill phase accounts for 30–40\% of the total inference latency. As finer-grained parallel decoding designs continue to reduce the workload of the decoding phase, the compute-bound prefill phase is expected to become increasingly dominant. This indicates the necessity of an arithmetic-intensity-aware inference framework that jointly addresses both compute and memory boundaries.

\begin{figure}[th]
    \centering    
    \subfloat[][]{
	\includegraphics[width=0.48\linewidth]{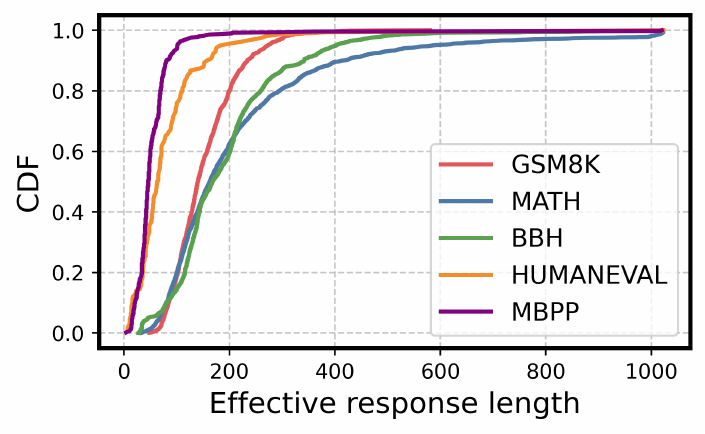}
        \label{fig:5a}
    }
    \subfloat[][]  {
	\includegraphics[width=0.48\linewidth]{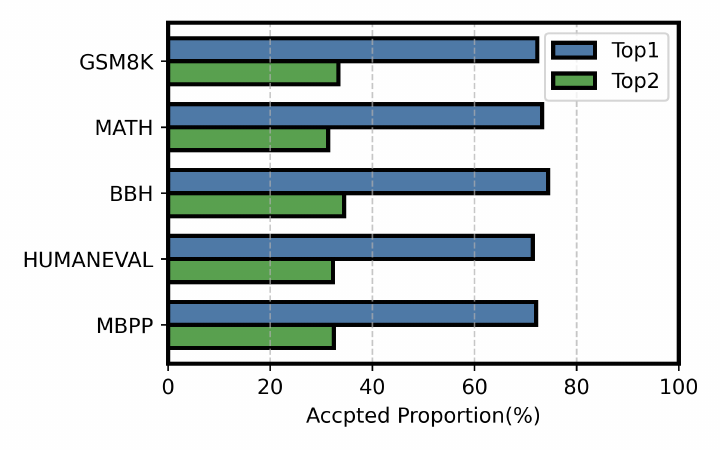}
        \label{fig:5b}
    }
    \caption{(a) Distribution of the effective response lengths and (b) Analysis of next-step acceptance rates for tokens that were initially unaccepted (with top-2 confidence).}
    \label{fig:5}
\end{figure}

\textbf{Observation 2: Previous dLLMs are unaware of the actual tasks, leading to excessively long response length.} Unlike conventional autoregressive LLMs, which decode sequentially until the $\texttt{[EOS]}$ token, each block in dLLMs leverages bidirectional attention, conditioning on both previous and subsequent tokens. Consequently, a fixed response length is needed to control the remaining output space. To ensure sufficient capacity, this length is typically set to a large default value (e.g., 1024), which 
introduces substantial computational overhead in the \textbf{compute-bound} prefill phase and results in redundant inference.
As illustrated in Figure~\figref{fig:5a}, our analysis of actual effective response lengths across multiple datasets reveals that for most queries, only 1/4 to 1/2 of the pre-allocated tokens are needed to generate correct answers, with the remainder filled with meaningless or repetitive content. 
This redundant portion often exhibits high confidence, meaning that each block typically requires very few decoding steps to converge.
It exacerbates the bottleneck, as the computationally expensive prefill phase becomes the dominant contributor to overall latency.

\textbf{Observation 3: A hard threshold delays the acceptance of potentially correct tokens, resulting in a longer decoding trajectory and more iterations.} During the \textbf{memory-bound} decoding phase, all masked tokens within a block are processed in parallel, but only those exceeding a confidence threshold are accepted, while others are re-masked. As shown in Figure~\figref{fig:5b}, a significant number of correct tokens are deferred simply because their confidence scores fall below the fixed threshold. Crucially, many of these tokens are then accepted as correct in subsequent rounds, leading to extra, unnecessary decoding steps and increased latency. Also, naively lowering the threshold compromises accuracy.

This observation presents a \textbf{clear optimization opportunity}: if we could speculatively accept those "actually correct" but below-threshold tokens, we could potentially terminate the decoding process much earlier. 
Extending speculative decoding to dLLM inference is non-trivial due to the model’s bidirectional attention, replacing the strictly ordered token tree used in conventional speculative decoding with a fully connected token graph. Moreover, modifying any single token requires recomputing an entire block, causing the computational cost of vanilla speculative decoding to grow rapidly with the number of candidates. This can render the process compute-bound, undermining its original purpose. Motivated by these mismatches, we develop a dLLM-specific speculative decoding framework, which will be described in \secref{subsec:Jump-share Speculative}.
\section{\method~Framework}
\label{sec:method}

Based on the insights from above analysis, we propose~\method, a dLLM inference framework that explicitly orchestrates the interleaved compute- and memory-bound phases. To mitigate the substantial prefill overhead inherent to dLLM inference, we introduce an adaptive length prediction strategy that reduces redundant computation (\secref{subsec:Adaptive Length Prediction}). In addition, \method~incorporates a dLLM-specific jump-share speculative framework (\secref{subsec:Jump-share Speculative}), which reduces decoding steps under limited arithmetic intensity.


\subsection{Adaptive Length Prediction}
\label{subsec:Adaptive Length Prediction}
To address the inefficiency of the prefill phase caused by fixed, predefined response lengths in dLLM inference, we propose an adaptive length prediction strategy. During the prefill phase, all tokens in the response sequence are computed and the KV cache is updated, providing a global decoding attempt. Although unmasking decisions are restricted to the current decoding block and all subsequent tokens are re-masked, these masked portions effectively capture the model’s inference draft. In particular, the positions of $\texttt{[EOS]}$ tokens implicitly encode the model’s intrinsic assessment of the required response length at the current inference step.

\begin{figure}[th]
    \centering
    \includegraphics[width=0.95\linewidth]{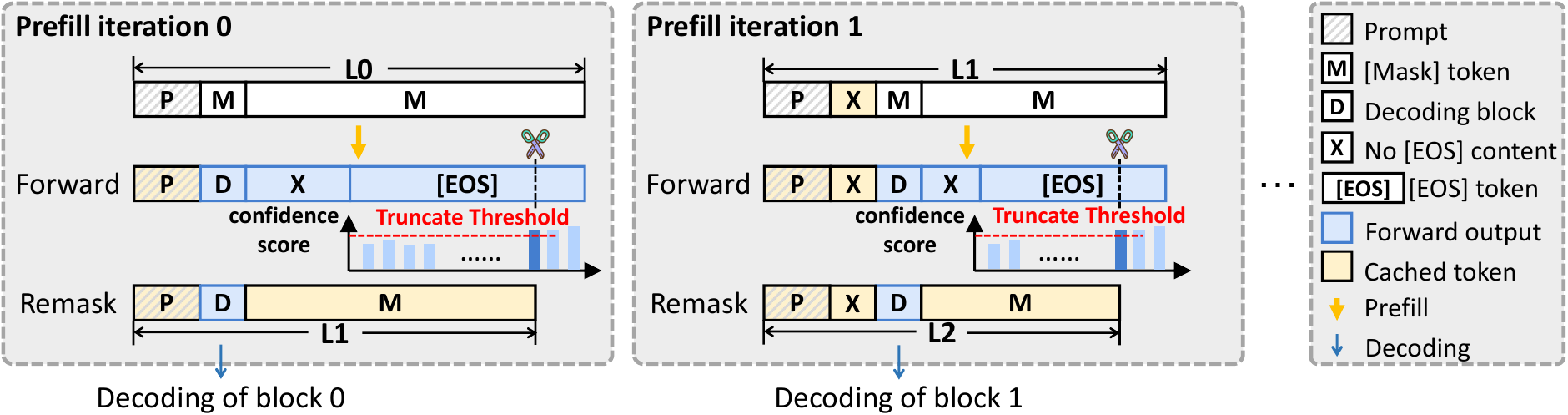}
    \caption{Adaptive length prediction strategy.}
    \label{fig:6}
\end{figure}

Accordingly, we monitor the entire response sequence before re-masking to detect any $\texttt{[EOS]}$ tokens during each prefill phase, as shown in~\figref{fig:6}. Once an $\texttt{[EOS]}$ token is identified, its confidence score is extracted and compared against a predefined truncate threshold. If the confidence exceeds the threshold, the response is truncated at that position. If no $\texttt{[EOS]}$ token is detected, or all detected tokens fall below the threshold, the response length remains unchanged. 

As inference progresses, the model’s uncertainty and diversity gradually converge, and our length prediction incrementally aligns with the actual required output. Compared to prior approaches such as DAEDAL which rely on an aggressive initialization followed by forced length expansion, or simply waiting for generation to reach $\texttt{[EOS]}$, our progressive strategy minimizes the prefill phase's arithmetic intensity and redundant computation while delivering measurable performance gains, which will be discussed in~\secref{sec:experiments}.

\subsection{Jump-share Speculative Decoding}
\label{subsec:Jump-share Speculative}
\begin{figure}[th]
    \centering
    \includegraphics[width=0.95\linewidth]{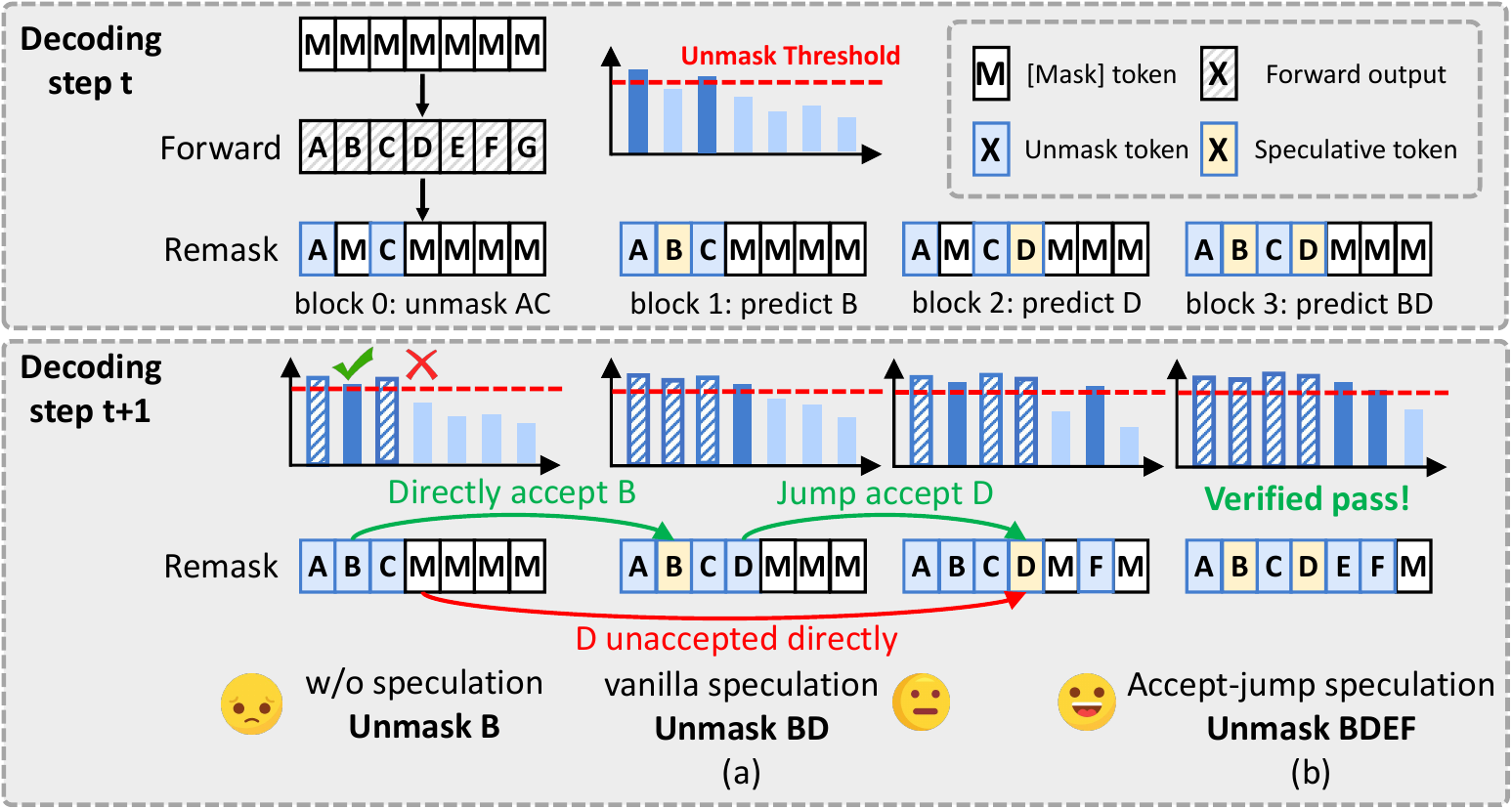}
    \caption{Inference process of (a) traditional dLLM speculative and (b) accept-jump speculative decoding.}
    \label{fig:7}
\end{figure}

To further enhance inference efficiency, we integrate speculative sampling into the memory-bound parallel decoding phase to increase the token acceptance rate. Specifically, among the tokens whose confidence scores fall below the predefined acceptance threshold, we select the top-k tokens as speculative candidates. Due to the bidirectional attention of dLLMs, modifying any token within a block affects the attention distribution and decoding behavior of all other tokens. Consequently, each speculative attempt, even one involving the unmasking of a single additional token, requires constructing an independent block. Similar to Spiffy\cite{agrawal2025spiffy}, we append multiple speculative blocks to the original response sequence, each of which represents a candidate decoding path, as illustrated in~\figref{fig:7}. We also design a block-wise attention mask that allows the decoding block and speculative blocks to access the globally cached context while remaining mutually invisible to each other.

However, this naive speculative framework is insufficient for dLLM inference. While speculative decoding benefits from more candidate predictions to improve the acceptance rate, the arithmetic intensity in dLLMs scales linearly with the number of speculative blocks. Consequently, the arithmetic intensity can quickly exceed the memory-bound limit of hardware platforms, shifting inference back to a compute-bound regime. Prior works such as Spiffy\cite{agrawal2025spiffy} evaluated performance gains primarily in terms of reduced decoding steps, achieving only marginal acceleration in practice. To overcome this, we propose accept-jump speculative and decoded-share speculative strategies, which collaboratively improve the token acceptance rate in two stages while strictly maintaining arithmetic intensity within the memory-bound region.

\subsubsection{Accept-jump Speculative Strategy}

Conventional speculative decoding for autoregressive models relies on a lightweight draft model to predict future tokens, constructing a fixed token tree. 
Under causal attention, decoding remains strictly sequential: acceptance halts at the first mismatch, and all non-surviving branches are discarded. 
In contrast, as discussed in \secref{sec:motivation}, the bidirectional attention in dLLMs removes these ordering constraints. Instead, multiple candidates can form a fully connected token graph, where any independently verified pair of nodes unlocks their connecting edge, enabling more flexible speculative exploration.

As shown in~\figref{fig:7}, we denote the main decoding block as $block_{0}$ and construct only three additional speculative blocks $block_{1}$, $block_{2}$, and $block_{3}$ to remain within the arithmetic intensity constraint. These speculative blocks correspond to accepting the top-confidence token ($B$), the second-highest-confidence token ($D$), and both tokens ($B$, $D$), respectively. 

During the next decoding step, we first examine whether $B$ and $D$ are unmasked in $block_{0}$.
If both of them are unmasked, we jump directly to $block_{3}$ and incorporate its decoding. 
Otherwise, in case that only one of them is unmasked---for instance, the token $B$---we proceed to $block_1$ where $B$ has been speculatively unmasked in advance, and verify whether the token $D$ can subsequently be accepted. If this verification succeeds, we jump again to $block_{3}$ and incorporate its decoding; if it fails, we directly adopt the decoding result produced by $block_1$. With this jump-accept speculative mechanism, more tokens can be unmasked within a single decoding iteration, substantially improving decoding efficiency under a limited candidate budget.

\subsubsection{Decoded-share Speculative Strategy}
While the accept-jump speculative strategy mitigates the arithmetic-intensity constraints on speculative diversity, its token acceptance rate remains limited due to insufficient predictions, as it considers only two additional tokens. Inspired by Fast-dLLM’s inter-block KV refresh and intra-block static KV caching, we propose a key insight: since the KV of tokens outside the current decoding block can be reused across multiple decoding iterations within a single block cycle, it is natural to ask whether the KV of already decoded tokens within a block can also be shared among multiple highly similar speculative blocks.

\begin{figure}[th]
    \centering
    \includegraphics[width=0.75\linewidth]{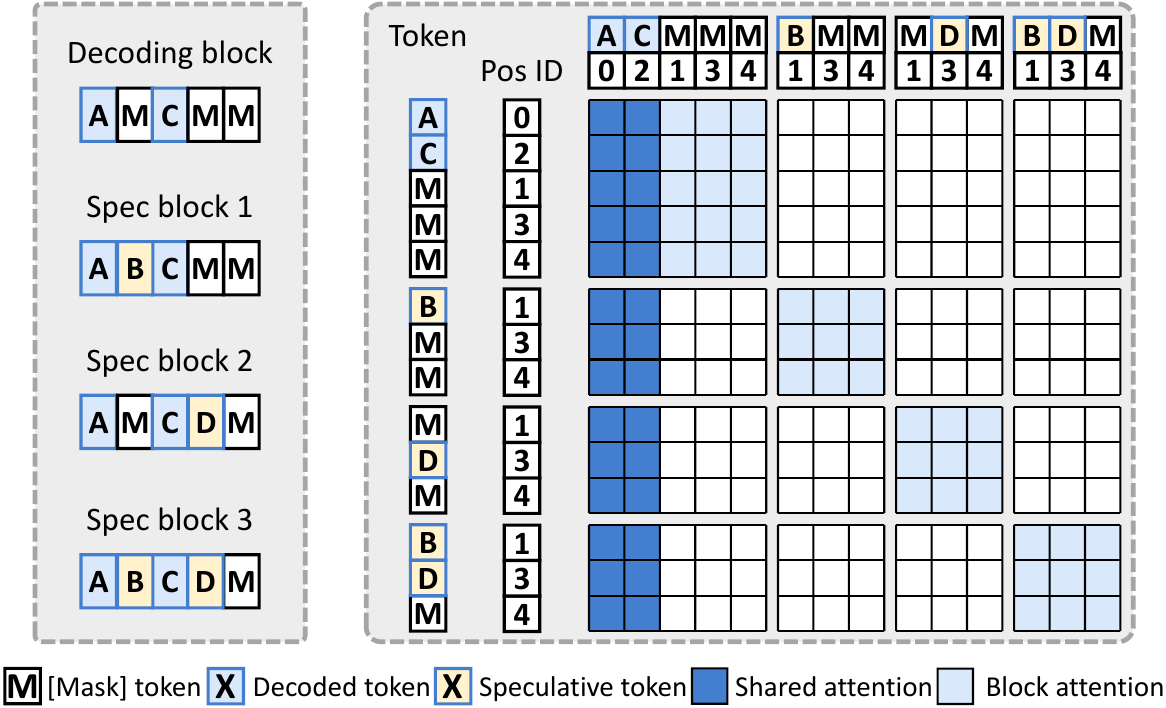}
    \caption{Position IDs and attention mask with decoded-share speculative strategy. For clarity of illustration, the figure depicts four blocks of five tokens each, rather than the eight blocks of 32 tokens used in actual dLLM inference.}
    \label{fig:8}
\end{figure}

As illustrated in~\figref{fig:8}, our speculative framework operates in two stages. In the first stage, when only a small number of tokens have been decoded, we employ the accept-jump mode, performing full computation for each speculative block to maximize accuracy. As decoding progresses and more tokens are unmasked, the second stage extracts already decoded tokens and recomputes them only once in $block_{0}$, while sharing their KV pairs across all speculative blocks. This approach progressively reduces the computational cost of speculative blocks as decoding advances, creating capacity to incorporate additional candidates. To simplify control logic and achieve higher end-to-end speedup, we fix the number of extra candidates to four: the third-highest confidence token, the top three tokens, the fourth-highest confidence token, and the top four tokens that fall below the acceptance threshold.
\section{Experimental Results}
\label{sec:experiments}

\subsection{Experimental Setup}
We evaluate~\method~on two representative open-source diffusion language models, LLaDA-Instruct \cite{nie2025large} and LLaDA-1.5 \cite{zhu2025llada15}. To assess the generality of our approach, our evaluation spans five benchmarks with lm-eval library: GSM8K \cite{cobbe2021training}, MATH \cite{hendrycks2021measuring}, BBH \cite{suzgun2023challenging}, HUMANEVAL \cite{chen2023accelerating}, and MBPP \cite{austin2021program}, covering a variety of tasks including mathematics, reasoning, and programming. Our framework is implemented on top of Fast-dLLM, inheriting its support for parallel decoding and Dualcache mechanisms. We disable padding and batch inference to maintain strict consistency with the vanilla LLaDA baseline. All experiments are performed on an NVIDIA A100 80GB GPU.

\subsection{Main Results \& Ablation Study}

\begin{figure}[th]
    \centering
    \begin{minipage}{0.48\linewidth}
        \centering
        \includegraphics[width=\linewidth]{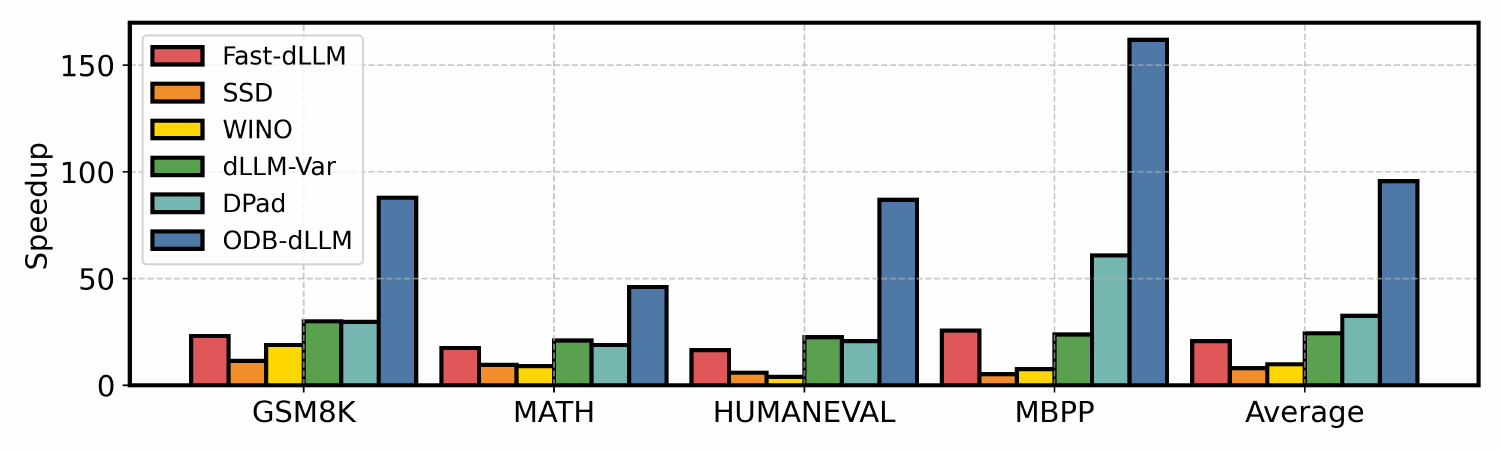}
        \caption{Comparison of~\method~performance against prior works.}
        \label{fig:9}
    \end{minipage}
    \hfill  
    \begin{minipage}{0.48\linewidth}
        \centering
        \includegraphics[width=\linewidth]{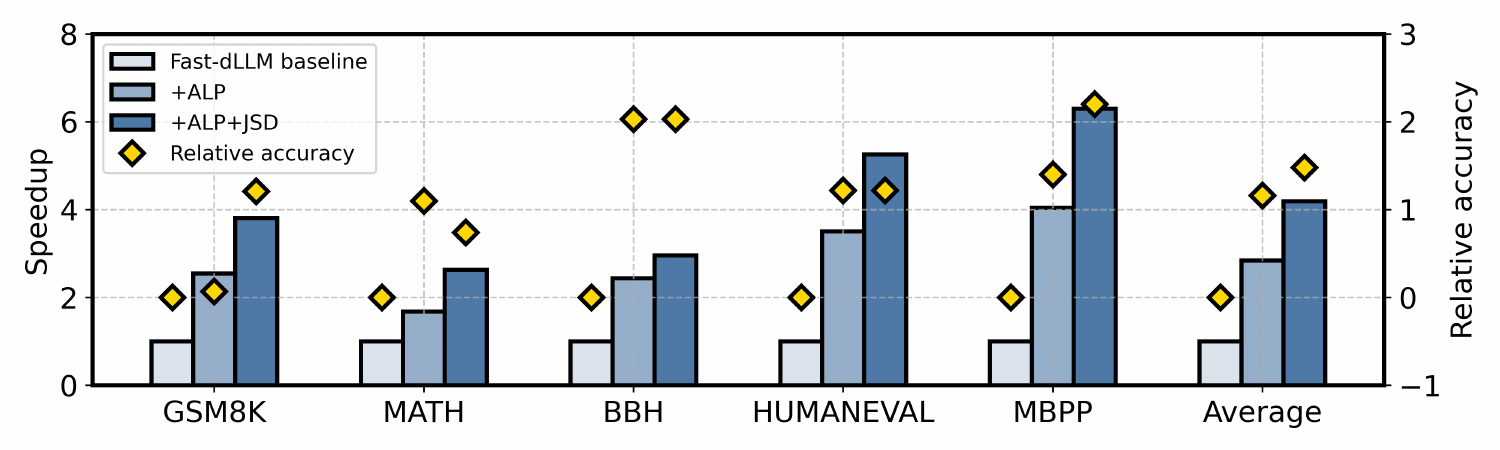}
        \caption{Ablation study on the speedup and accuracy relative to Fast-dLLM.}
        \label{fig:10}
    \end{minipage}
\end{figure}

We progressively augment the vanilla models with Fast-dLLM and our~\method~framework, as summarized in~\tabref{tab2} and~\tabref{tab3}. Relative to the vanilla baselines, \method~achieves 46×–162× speedup on LLaDA-Instruct and 50–182× speedup on LLaDA-1.5 across multiple datasets. Compared with Fast-dLLM, \method~also delivers 2.63–6.30× and 2.60–7.22× speedup on the two models, respectively. Notably, our framework substantially mitigates the accuracy degradation introduced by parallel decoding and approximate KV caching in Fast-dLLM, yielding performance that matches or even surpasses the vanilla models. We additionally compare against recent dLLM acceleration techniques, including WINO \cite{hong2025wide} and SSD \cite{gao2025self}, which also employ speculative strategies, as well as dLLM-Var \cite{yang2025diffusion} and DPad \cite{chen2025dpad}, which handle the predefined response length. As shown in~\figref{fig:9}, \method~consistently achieves the highest speedup across all baselines.

\begin{table}[!th]
\caption{Accuracy and inference speed across diverse benchmarks on LLaDA-Instruct}
\centering
\small
\begin{tabular}{cc|ccc}
\toprule
Benchmark & Metric & Vanilla & +Fast-dLLM & +\method\\
\midrule
GSM8K & Acc. & 79.76 & 77.79 & 79.00 \\
(\textit{5-shot}) & Speed & 1.00/-- & 23.08/1.00 & 87.96/3.81  \\
\midrule
MATH & Acc. & 31.56 & 29.82 & 30.92 \\
(\textit{4-shot}) & Speed & 1.00/-- & 17.54/1.00 & 46.11/2.63  \\
\midrule
BBH & Acc. & 55.98 & 53.20 & 55.23 \\
(\textit{3-shot}) & Speed & 1.00/-- & 23.97/1.00 & 70.86/2.96  \\
\midrule
HUMANEVAL & Acc. & 39.63 & 35.98 & 37.20 \\
(\textit{0-shot}) & Speed & 1.00/-- & 16.54/1.00 & 86.96/5.26  \\
\midrule
MBPP & Acc. & 37.80 & 33.00 & 35.20 \\
(\textit{3-shot}) & Speed & 1.00/-- & 25.70/1.00 & 161.79/6.30  \\
\bottomrule
\end{tabular}
\label{tab2}
\end{table}

\begin{table}[!th]
\caption{Accuracy and inference speed across diverse benchmarks on LLaDA-1.5}
\centering
\small
\begin{tabular}{cc|ccc}
\toprule
Benchmark & Metric & Vanilla & +Fast-dLLM & +\method\\
\midrule
GSM8K & Acc. & 81.80 & 80.97 & 81.05 \\
(\textit{5-shot}) & Speed & 1.00/-- & 22.50/1.00 & 95.10/4.23  \\
\midrule
MATH & Acc. & 33.26   & 31.66 & 31.96 \\
(\textit{4-shot}) & Speed & 1.00/-- & 19.33/1.00 & 50.26/2.60  \\
\midrule
BBH & Acc. & 56.89 & 54.14 & 55.74 \\
(\textit{3-shot}) & Speed & 1.00/-- & 23.05/1.00 & 78.35/3.40  \\
\midrule
HUMANEVAL & Acc. & 37.20 & 38.41 & 39.02 \\
(\textit{0-shot}) & Speed & 1.00/-- & 15.07/1.00 & 91.46/6.07  \\
\midrule
MBPP & Acc. & 39.00 & 33.80 & 34.80 \\
(\textit{3-shot}) & Speed & 1.00/-- & 25.23/1.00 & 182.14/7.22  \\
\bottomrule 
\end{tabular}
\label{tab3}
\end{table}


To analyze the dual benefits of accuracy and efficiency provided by~\method~over Fast-dLLM, we conduct an ablation study on LLaDA-Instruct, as shown in~\figref{fig:10}. With Adaptive Length Prediction (ALP) enabled, \method~achieves an average 2.84× speedup. This improvement stems from two main factors. First, ALP compresses the response length to an average of 272 tokens, substantially shorter than the original 1024 and the 625 in DAEDAL  \cite{li2025beyond}. This eliminates significant redundancy in the prefill-dominant region following the $\texttt{[EOS]}$ token, as discussed in~\secref{sec:motivation}. Second, after accounting for the prompt length, we measure reductions in per-block prefill length and latency (\tabref{tab4}), confirming meaningful arithmetic intensity savings in the compute-bound phase. Importantly, the progressive contraction of response length during each prefill iteration implicitly constrains the model to a more accurate answer space, producing higher accuracy than simply truncating after the first generated $\texttt{[EOS]}$ and yielding a 1.16-point improvement in overall performance.


\begin{table}[!th]
\caption{Average per-block prefill length and prefill time reductions achieved by~\method.}
\centering
\small
\begin{tabular}{c|ccccc}
\toprule 
\multirow{2}{*}{Benchmark} & \multicolumn{2}{c}{Vanilla/Fast-dLLM} & DAEDAL & \multicolumn{2}{c}{\method} \\
\cmidrule(r){2-3} \cmidrule(lr){4-4} \cmidrule(l){5-6}
 & Length & Time(s) & Length & Length & Time(s) \\
\midrule
GSM8K(\textit{5-shot}) & 2031 & 0.189 & 1370 & 1247(61\%) & 0.149(79\%) \\
MATH(\textit{4-shot}) & 1709 & 0.163 & 1389 & 1016(59\%) & 0.123(75\%) \\
BBH(\textit{3-shot}) & 1988 & 0.182 & -- & 1244(63\%) & 0.138(76\%) \\
HUMANEVAL(\textit{0-shot}) & 1170 & 0.128 & 959 & 472(40\%) & 0.056(44\%) \\
MBPP(\textit{3-shot}) & 1762 & 0.170 & 1356 & 912(52\%) & 0.114(67 \%) \\
\bottomrule 
\end{tabular}
\label{tab4}
\end{table}

With Jump-Share Speculative Decoding(JSD), we observe even greater gains, achieving a 4.19× speedup over Fast-dLLM along with a 1.48-point improvement in accuracy. Unlike speculative decoding for autoregressive LLMs, which does not affect the target model’s final output, speculative decoding in dLLMs can alter the token acceptance order and response content due to bidirectional attention and parallel generation. A detailed analysis of this phenomenon is provided in the following subsection.


\subsection{Speculative Variants}

\begin{figure}[th]
    \centering    
    \subfloat[][]{
	\includegraphics[width=0.48\linewidth]{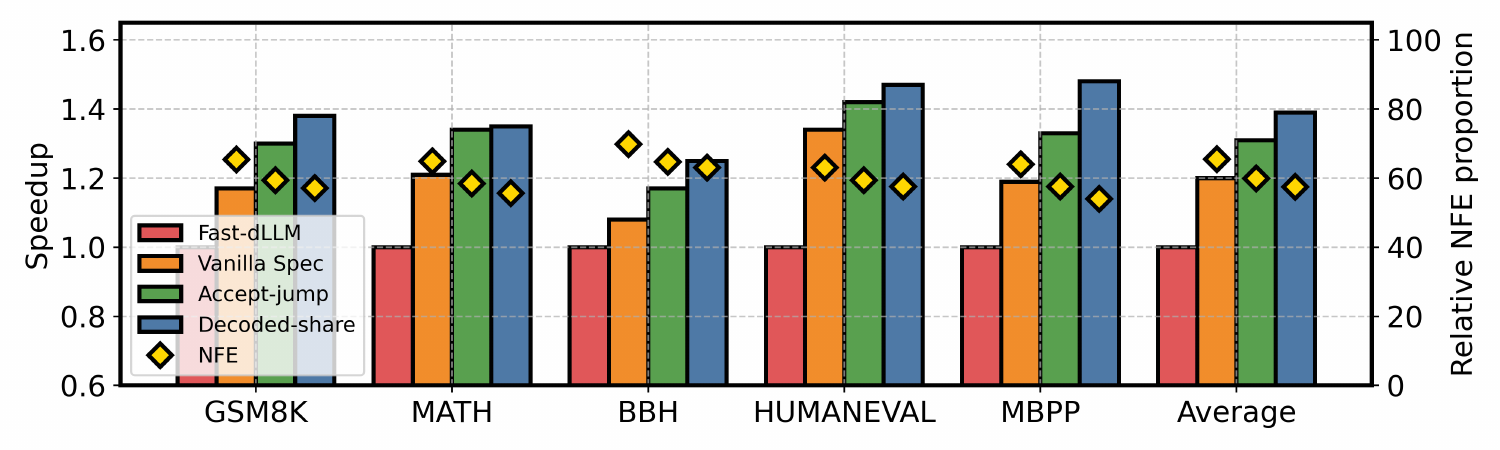}
        \label{fig:11a}
    }
    \subfloat[][]  {
	\includegraphics[width=0.48\linewidth]{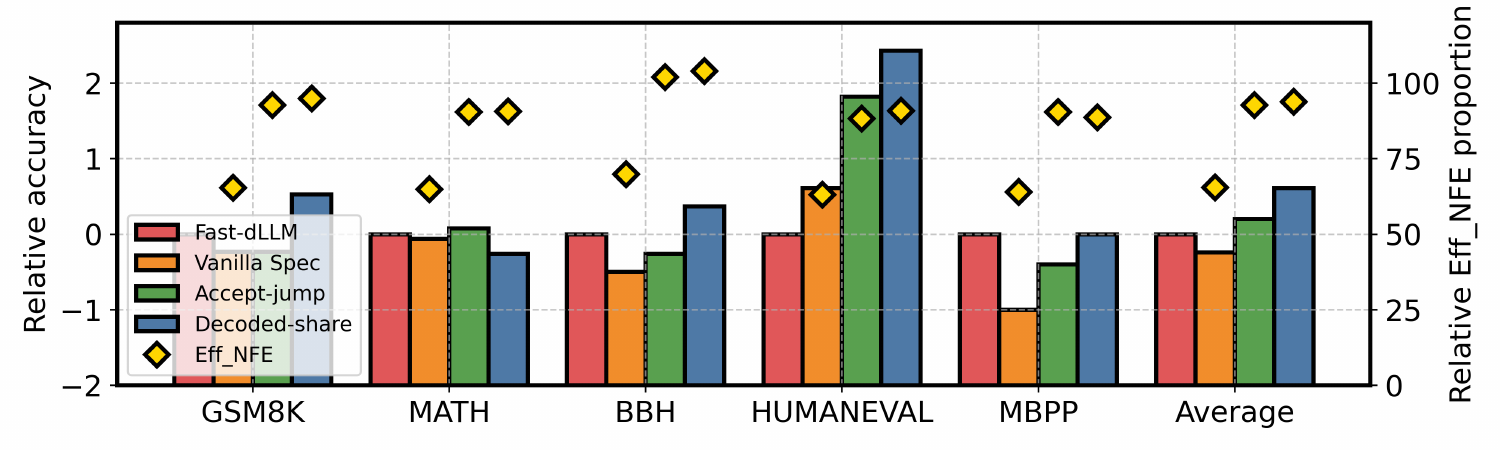}
        \label{fig:11b}
    }
    \caption{Comparison of (a) relative NFE proportion and speedup and (b) relative effective NFE proportion and accuracy on different speculative variants.}
    \label{fig:11}
\end{figure}

To isolate the accuracy-efficiency trade-offs of different speculative variants, we apply speculative decoding to Fast-dLLM without length prediction. As shown in~\figref{fig:11a}, vanilla speculation offers limited speedup despite fewer number of function evaluations (NFEs), mainly due to the small candidate set and an attention mask incompatible with Flash Attention. Our accept-jump strategy increases the number of accepted tokens per step, delivering substantially higher speedup combined with memory-efficient attention. The introduction of decoded-share speculation further reduces the number of decoding rounds, yielding an additional non-trivial speedup.

Since speculative decoding in dLLMs alters the decoding trajectory, it is essential to analyze its impact on model accuracy. As shown in~\figref{fig:11b}, vanilla speculative decoding exhibits notable accuracy degradation. To quantify how decoding-state transitions affect final correctness, we introduce the effective NFE (Eff\_NFE) metric, which measures the number of effective state changes in the unmasking sequence. Tokens unmasked within the same step are sampled independently, and bidirectional attention can induce potential conflicts among these tokens. A lower Eff\_NFE indicates a higher likelihood of such conflicts. For vanilla speculative decoding, Eff\_NFE equals NFE, and its reduction inevitably harms accuracy. In contrast, accept-jump speculative induces multiple trajectory jumps within a single step, resulting in a higher Eff\_NFE and improved accuracy, even surpassing the non-speculative baseline. Furthermore, while decoded-share speculative produces only modest gains in Eff\_NFE, it preferentially unmasks tokens that previously failed acceptance yet consistently exhibit high confidence. This behavior implicitly reinforces the model’s stable preferences for these tokens, resulting in the highest overall accuracy among all variants.


\section{Conclusion}
In this paper, we propose~\method, an arithmetic intensity inspired framework for accelerating diffusion-based large language models. By analyzing the interleaved compute- and memory-bound phases in existing dLLM inference frameworks, \method~introduces adaptive length prediction strategy and jump-share speculative decoding to optimize computation-memory characteristics on hardware platforms, thereby maximizing inference efficiency. Extensive experiments across multiple datasets show that~\method~achieves 46-162× and 2.63-6.30× speedup over baseline dLLMs and Fast-dLLM, respectively, while effectively mitigating accuracy degradation.

\clearpage
{
\small
\bibliographystyle{cite}
\bibliography{cite}
}

\end{document}